\documentclass[11pt]{article}

\usepackage[preprint]{acl}

\usepackage{times}
\usepackage{latexsym}

\usepackage[T1]{fontenc}

\usepackage[utf8]{inputenc}

\usepackage{microtype}

\usepackage{inconsolata}

\usepackage{graphicx}
\usepackage{subcaption}
\usepackage{amsfonts}
\usepackage{caption}
\usepackage{colortbl}
\usepackage{color}
\usepackage[most]{tcolorbox} 
\tcbuselibrary{breakable}
\usepackage{multirow}
\usepackage{multicol}
\usepackage{adjustbox}
\usepackage{tabularx}
\usepackage{enumitem}
\usepackage{xspace}
\usepackage{booktabs}
\usepackage{array}
\usepackage{longtable}
\usepackage{url}
\usepackage{threeparttable}

\definecolor{paleviolet}{HTML}{F4EEFF}

\newcommand{\methodabbr}{SAGE}           
\newcommand{\method}{\textsc{\methodabbr}\xspace}  

\title{SAGE: Sparse Adaptive Guidance for Dependency-Aware\\Tabular Data Generation}

\author{Shuo Yang$^*$ \; Zheyu Zhang$^*$ \; Bardh Prenkaj \;  
         Gjergji Kasneci \vspace{0.3cm}\\
    Technical University of Munich \\
    Munich Center for Machine Learning (MCML) \\
{\small \tt \{name.surname\}@tum.de}}

\begin{document}
\maketitle

\def\thefootnote{*}\footnotetext{Equal contribution.}\def\thefootnote{\arabic{footnote}}

\begin{abstract}
Generating high-fidelity synthetic tabular data remains a critical challenge for enhancing data availability in privacy-sensitive and low-resource domains. Recent approaches leverage LLMs by representing table rows as sequences, yet suffer from two fundamental limitations: (1) they model feature dependencies densely, introducing spurious correlations; and (2) they assume static relationships between features, ignoring how these dependencies vary with feature values. To overcome these limitations, we introduce SAGE (Sparse Adaptive Guidance), a novel LLM-based generation framework that enforces sparse and dynamic dependency guidance. SAGE discretizes features into value-aware pseudo-features and constructs a mutual information-based sparse dependency graph. This graph adaptively guides generation through explicit context selection or implicit logit correction, enabling LLMs to focus on truly relevant information during synthesis. Our extensive experiments across six datasets and multiple tasks reveal that SAGE not only improves data fidelity and downstream utility, boosting F1 scores by 10\% compared to previous LLM-based methods, but also reduces policy violations by one point. These results highlight the importance of adaptive structure in tabular data generation and provide new insights into context-sensitive control of LLMs.\footnote{Our code is publicly available at \url{https://github.com/ShuoYangtum/SAGE}.}
\end{abstract}

\section{Introduction}

Tabular data forms the backbone of decision-making across healthcare~\citep{healthcare}, finance~\citep{finance}, and education~\citep{education}, yet obtaining high-quality datasets remains challenging due to privacy constraints and data collection costs~\cite{borisov2022deep}. This scarcity has driven significant interest in synthetic tabular data generation, which promises to unlock data-driven applications while preserving privacy~\citep{liu2024tabular, zhao2025tabula}.

\begin{figure}[!t]
    \centering
       \includegraphics[width=.9\linewidth]{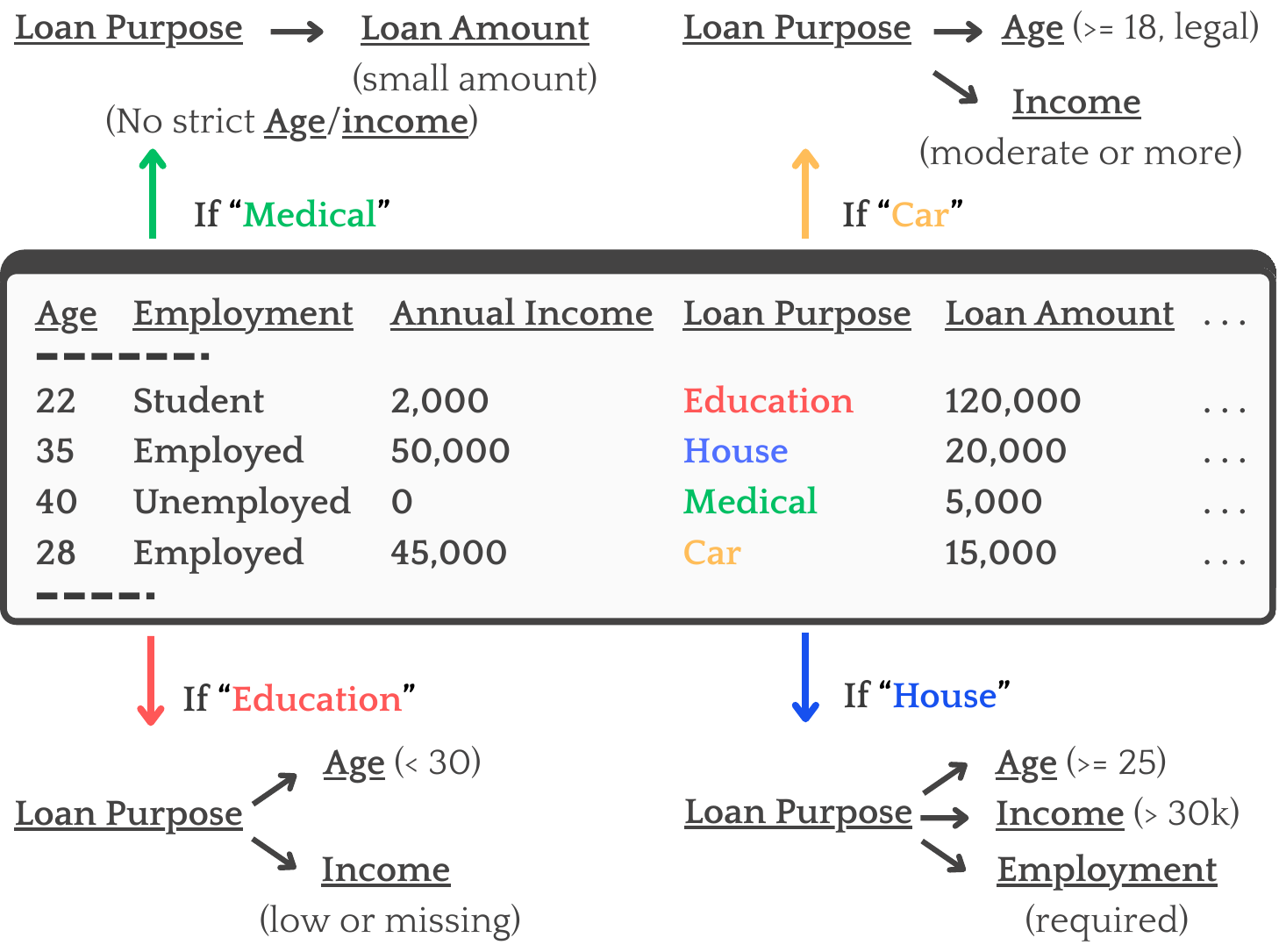}
    \caption{Value-conditioned dynamic dependencies in tabular data. Dependencies such as \textit{Loan Purpose $\rightarrow$ Age, Income} vary with purpose: ``Education'' loans imply youth and low income, while ``House'' loans require stable employment and higher income. Static models overlook such conditional structures.}
    \label{fig:example}
    \vspace{-10pt}
\end{figure}

Early approaches have primarily focused on learning underlying data distributions through neural generative models~\citep{stoian2025survey}. Traditional methods based on variational autoencoders~\citep{xu2019modeling} and generative adversarial networks~\citep{Kamthe2021CopulaFF} learn tabular value distributions and generate new records by sampling from the latent space.
However, these methods often produce logically inconsistent records, such as pairing professional occupations with minor ages~\citep{yang-etal-2024-p, long2025llm}. 

Recent work has addressed this limitation by leveraging large language models (LLMs), which benefit from strong sequential modeling capabilities and rich pre-trained knowledge~\citep{liu2023pre, zhang-etal-2023-babys, han-etal-2025-attributes}. These approaches convert tabular rows to textual sequences using templates like ``feature is value''~\citep{borisov2022language}, naturally incorporating world knowledge and dramatically reducing implausible outputs.
However, current LLM-based generators face two fundamental limitations. First, they model feature relationships densely through fully-connected attention mechanisms, despite tabular data being inherently sparse~\citep{liu2023goggle}, with dependencies present only among limited subsets of features. This introduces spurious correlations and computational overhead. Second, they assume static dependencies between features, failing to capture how relationships change with specific values. Figure~\ref{fig:example} illustrates this dynamic behavior: ``Education'' loans correlate with young applicants and low income, while ``House'' loans require stable employment and higher income thresholds. Existing dependency-aware methods~\citep{xu2024llms} rely on pre-annotated, fixed dependency graphs that cannot adapt to such value-conditioned patterns.

As a two-pronged solution, we propose \method (Sparse Adaptive Guidance), a novel framework that enforces both sparse and dynamic dependency modeling for LLM-based tabular data generation. \method discretizes features into value-aware pseudo-features and constructs mutual information-based sparse dependency graphs that dynamically adapt during generation. This enables the model to focus on truly relevant feature relationships while filtering out spurious correlations. We propose two guidance strategies: (1) explicit context selection through \emph{Feature Selector} and (2) implicit adjustment via \emph{Logit Correction}, both designed to ensure that LLMs condition on contextually appropriate information when generating each feature value.

\noindent Our contributions are summarized as follows:
\begin{enumerate}[leftmargin=*, nolistsep]
    \vspace{3pt}
    \item We propose \method, a novel tabular data synthesis framework that jointly models sparse feature dependencies and their dynamic variations based on feature values. Unlike existing methods that assume static relationships, \method adapts dependency structures during generation through mutual information-guided pseudo-feature discretization.
    
    \vspace{3pt}
    \item We introduce two complementary guidance mechanisms: \textit{Feature Selector} for explicit context filtering and \textit{Logit Correction} for implicit confidence adjustment. Additionally, we present engineering optimizations including value-only loss computation and rejection sampling to address computational overhead and invalid value generation in LLM-based approaches.
    
    \vspace{3pt}
    \item We conduct comprehensive experiments across six diverse datasets, evaluating downstream utility, data fidelity, privacy preservation, and realism. SAGE consistently outperforms existing methods, achieving up to 10\% F1 improvement on classification tasks and reducing policy violation rates by over 1 percentage point compared to state-of-the-art LLM-based generators.
\end{enumerate}

\section{Related Work}

\noindent\textbf{End-to-End Tabular Generative Models.}
Tabular data generation methods rely on end-to-end neural architectures to capture the full joint distribution \citep{hollmann2025accurate}. Unlike vision or NLP domains, tabular datasets are often limited in size, making synthetic data generation particularly valuable for data augmentation \citep{abs-2504-16506}. \citet{xu2019modeling} pioneered this field with CTGAN and TVAE, addressing imbalanced categorical features through adversarial training and providing stable probabilistic generation via variational autoencoders. Subsequent GAN-based methods \citep{9679177, kim2021oct, pmlr-v157-zhao21a, zhao2024ctab} refined adversarial training but often suffer from mode collapse and poor interpretability. More recently, diffusion-based models such as TabDDPM \citep{pmlr-v202-kotelnikov23a}, FinDiff \citep{finance}, AutoDiff \citep{abs-2310-15479}, and \textsc{TabSyn} \citep{zhang2023mixed} emerged as promising alternatives through iterative denoising processes and architectural improvements. However, these models treat data as value matrices, largely ignoring the semantic meaning of features.

\noindent\textbf{Language Models for Tabular Data Modeling.}
Motivated by the impressive performance of large language models, recent research explores tabular data generation by representing table rows as sequences of feature-value pairs. \textsc{GReaT} \citep{borisov2022language} first demonstrated this approach through autoregressive language modeling, effectively leveraging pretrained world knowledge. Subsequent works like Pred-LLM \citep{abs-2410-21717} and TabuLa \citep{zhao2025tabula} optimized feature-value representations to enhance correlation modeling, while P-TA \citep{yang-etal-2024-p} used proximal policy optimization to integrate GAN-based discriminator feedback. Prompt-based methods have emerged as an alternative paradigm: EPIC \citep{NEURIPS2024_37f2f382} and TabGen-ICL \citep{abs-2502-16414} demonstrated effective in-context learning for tabular synthesis, and CLLM \citep{seedat2023curated} leveraged LLM prior knowledge for data augmentation in low-data regimes. While these methods effectively incorporate feature semantics, they overlook the structured nature of feature dependencies by modeling entire rows as flat sequences, leading to high sampling latency.

\noindent\textbf{Dependency Modeling in Tabular Data.}
To address the limitations of black-box models and flat-sequence representations, another line of research explicitly models the sparse, structured dependencies in tabular data. Early approaches integrated structural priors into classical generative frameworks: GOGGLE \citep{liu2023goggle} encodes pairwise feature relationships into graphs within a VAE, GANBLR \citep{9679177} incorporates auxiliary Bayesian Networks into a GAN, and DRL \citep{stoian2025beyond} imposes logical rules through differentiable layers compatible with gradient-based training. More recently, structure-aware LLM-based methods constrain generation to follow predefined structures. \textsc{SPADA}~\citep{spada} induces a sparse dependency graph that dictates the generation process, PAFT \citep{xu2024llms} aligns generation with a statistically-determined feature order, and \textsc{GraDe}~\citep{grade} uses statistical dependencies to dynamically guide the language model's attention. However, these approaches share a critical assumption: the dependency structure is predetermined and static throughout generation. This static view is limiting, as it fails to capture context-dependent feature relationships where one feature's value can dynamically alter dependencies among others. Our work challenges this assumption by proposing an adaptive framework where dependency structures evolve dynamically during synthesis.

\section{Methodology}
\begin{figure*}[!t]
    \centering
    \includegraphics[width=\textwidth]{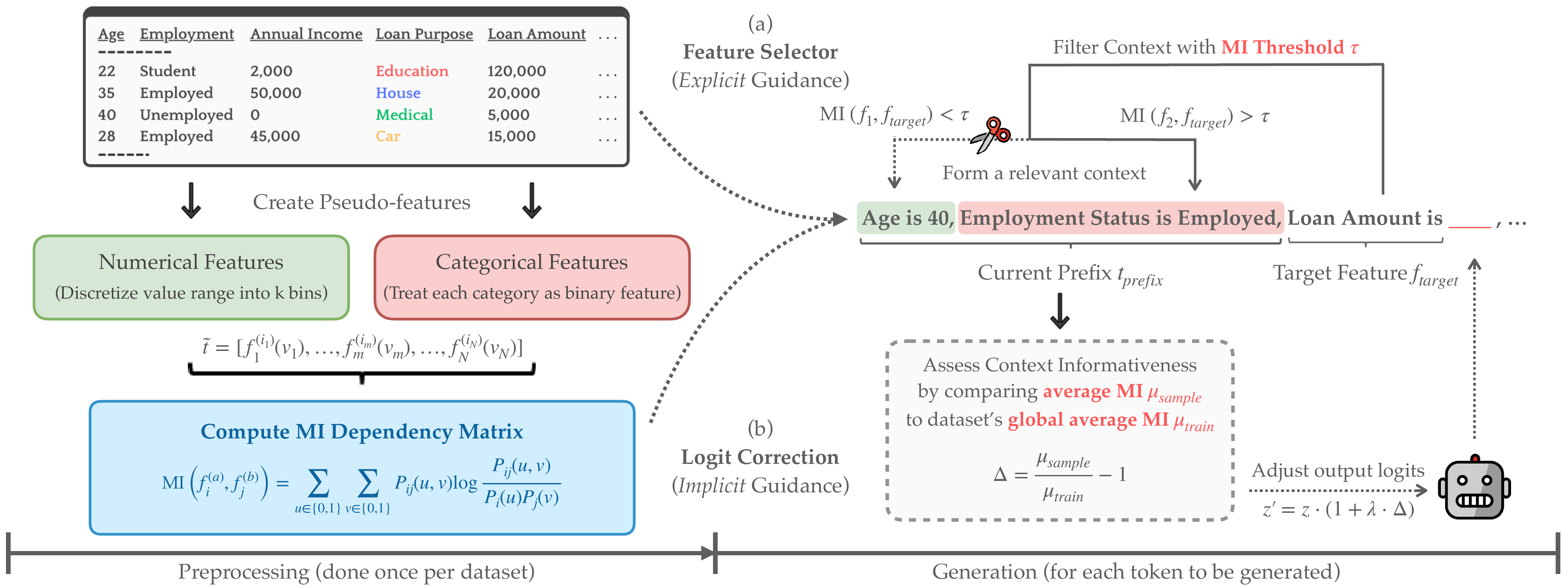}
    \caption{Overview of \method. In the preprocessing stage (\textbf{left}), a mutual-information-based dependency matrix is constructed from the data. During generation (\textbf{right}), this matrix guides the model using one of two strategies: (a) \textit{Feature Selector}, which provides explicit guidance by pruning the input context with an MI threshold $\tau$; and (b) \textit{Logit Correction}, which provides implicit guidance by adaptively adjusting the output logits according to the informativeness of the current context.}
    \label{fig:pipeline}
    \vspace{-10pt}
\end{figure*}
\subsection{Problem Formulation}

Let $\mathcal{T}$ be a tabular dataset with $N$ samples, $t_i \in \mathcal{T}$ is represented by $F$ features $(f_1,\dots, f_F)$. Each feature $f_j$ in sample $t_i$ takes a specific value $v_{ij}$. Following the standard taxonomy in tabular toolbox~\citep{SDV}, we categorize the features into two disjoint sets: $\mathcal{F}_{\text{num}}$ and $\mathcal{F}_{\text{cat}}$, such that the total set of features is $\mathcal{F}=\mathcal{F}_{\text{num}} \cup \mathcal{F}_{\text{cat}}$, where:
\begin{itemize} [leftmargin=*, nolistsep]
    \item $\mathcal{F}_{\text{num}}$: Continuous numerical variables, e.g. age.
    \item $\mathcal{F}_{\text{cat}}$: Discrete character variables with a finite set of possible values, e.g. marital status.
\end{itemize}%
Our objective is to learn the underlying distribution of the samples in $\mathcal{T}$ and to generate a new set of $M$ synthetic samples, denoted as $\hat{\mathcal{T}} = \{\hat{t}_1, \dots, \hat{t}_M\}$, where  $\hat{t} \notin \mathcal{T}$.

\subsection{Tabular Data Generator}
\subsubsection{Embedding Tabular Records to Text.}
To enable LLMs to process tabular data, we transform each sample $t_i$ into a natural language sentence $s_i$. Specifically, we convert structured records into sequences of \emph{<subject, predicate, object>} phrases, typically in the form of ``\textit{feature} is \textit{value}'' templates~\citep{borisov2022language}. Therefore, for each feature $f_j$ and its corresponding value $v_{ij}$ in sample $t_i$, we convert the pair into a short phrase of the form ``$f_j$ is $v_{ij}$''. All such phrases for a given sample are concatenated using commas to form the sentence $s_i$.
By applying the template, we obtain a set of textual representations $S = \{s_1, \dots, s_N\}$ corresponding to the original tabular dataset $\mathcal{T}$. We then fine-tune the LLM on $S$ to model the underlying data distribution.

\subsubsection{Training.}
Following the continued pretraining strategy adopted in \textsc{GReaT}~\citep{borisov2022language}, we fine-tune the LLM on $S$ by minimizing the negative log-likelihood of each target token of values $(v_{i1},...,v_{ij})$ in $s_i$, as shown in Eq.~\eqref{eq:generator}. 

\begin{equation}
\mathcal{L}_{\text{LM}}(\theta) = - \sum_{i=1}^{N} \sum_{t=1}^{|s^{v}_i|} \log P_\theta\left( s^{v}_{i,t} \mid s^{v}_{i,<t} \right),
\label{eq:generator}
\end{equation}%
where $s^{v}_{i,t}$ denotes the $t$-th value-related token in sentence $s_i$, and $\theta$ represents the model parameters.

To enhance the robustness of $\theta$ in modeling the feature-value distribution, we incorporate the permutation strategy from \textsc{GraDe}~\citep{grade}, where the order of ``\textit{feature is value}'' phrases in each $s_i$ is randomly shuffled during training. This discourages the model from learning spurious dependencies that arise solely from fixed sequence positions, which do not reflect true feature dependencies in $\mathcal{T}$. After that, the fine-tuned generator $\theta$ implicitly integrates the knowledge encoded in the pretrained LLM with the feature-value distribution learned from $\mathcal{T}$. This enables it to model inter-feature dependencies and reduces the risk of logical inconsistencies in generated samples.

Consistent with \textsc{GReaT} and \textsc{GraDe}, $\theta$ generates synthetic samples by randomly selecting a subset of real feature-value pairs as a prefix, and then autoregressively completing the remaining ones.
Formally, let $x_{1:k}$ denote the prefix consisting of $k$ tokens derived from a subset of feature-value pairs. The model then generates the remaining tokens $x_{k+1:L}$ by sampling from the conditional distribution:
\begin{equation}
P_\theta(x_{k+1:L} \mid x_{1:k}) = \prod_{t=k+1}^{L} P_\theta(x_t \mid x_{1:t-1}),
\end{equation}%
where $x_t$ denotes the $t$-th token in the generated sequence, and $L$ is the total length of the output. 

However, the attention mechanism forces each token to consider all prior tokens in $s_i$, even those unrelated to the current generation. As a result, $\theta$ can learn misleading dependencies between dependent features, leading to spurious correlations that degrade downstream model performance.

\subsection{Generation guided by Sparse and Dynamic Feature Dependency}
The core challenge lies in balancing sparsity and adaptability in dependency modeling. While existing methods either ignore feature relationships entirely or assume static connections, \method introduces a dynamic approach that adapts to feature values during generation.

Figure~\ref{fig:pipeline} illustrates the complete pipeline of our approach. \method operates in two phases: \textit{preprocessing} and \textit{generation}. During preprocessing, we discretize features into value-aware pseudo-features and construct a mutual information dependency matrix that captures statistical relationships between feature values. During generation, this matrix guides the LLM through two complementary strategies: (a) \emph{Feature Selector} provides explicit guidance by filtering the input context with a mutual information (MI) threshold~\citep{cover1999elements}, and (b) \emph{Logit Correction} offers implicit guidance by adaptively adjusting output logits based on the context's informativeness. This dynamic approach enables the model to focus on truly relevant dependencies while adapting to value-conditioned patterns during synthesis.

Methods leveraging predefined feature graphs~\citep{spada, grade} have demonstrated general superiority of modeling sparse dependencies over fully connected attention. However, we argue that these approaches fail to account for the dynamic nature of fine-grained feature-value dependencies during generation.

Formally, let $\mathcal{G} = (\mathcal{F}, \mathcal{E})$ represent a static feature dependency graph, where $\mathcal{F}$ denotes the set of features and $\mathcal{E} \subseteq \mathcal{F} \times \mathcal{F}$ encodes pairwise dependencies. However, when the value of a particular feature $f \in \mathcal{F}$ changes during generation, it may dynamically influence how other features relate to $f$, thereby modifying their relevance or conditional influence. As a result, this dynamic nature of feature interactions poses a challenge for approaches that rely on externally defined and static dependency graphs, as they struggle to adapt to evolving dependencies that emerge throughout the generative process. Consequently, static methods may suffer from reduced flexibility and accuracy, particularly in scenarios where the semantics of a given feature value significantly reshape the dependencies among remaining features.

As a solution, \method jointly accounts for the \textbf{sparsity} of feature dependencies and the \textbf{value-aware dynamics} of such dependencies during generation. To achieve this, we first discretize each feature $f$ into $k$ \emph{pseudo-features} based on its value domain, denoted as:
\begin{equation}
\text{Bin}(f) = \{f^{(1)}, f^{(2)}, \dots, f^{(k)}\},
\end{equation}%
where each $f^{(i)}$ represents a bin corresponding to a specific sub-range or category of values for $f$.

Let $\Delta_f=\max(f)-\min(f)$ and $w_f=\Delta_f/k$.  
Set the cut-points $a_i=\min(f)+i\,w_f$ for $0\!\le\! i\!\le\! k$.  
The $k$ binary bins are  
\begin{equation}
\begin{aligned}
f^{(i)}(v) &= \mathbb{I}[\,a_{i-1}\le v<a_i\,],  &(1\le i<k),\\
f^{(k)}(v) &= \mathbb{I}[\,a_{k-1}\le v\le a_k\,].
\end{aligned}
\end{equation}%
Half-open intervals keep bins disjoint, and the right-closed last bin
captures $v=\max(f)$. The number of bins $k$ is set by the Freedman-Diaconis rule
$k=\bigl\lceil\Delta_f/(2\,\mathrm{IQR}(f)\,n^{-1/3})\bigr\rceil$,
capped at $k\le16$ to control sparsity. 

For categorical features $f \in \mathcal{F}_{\text{cat}}$, we treat each possible category $c \in \mathcal{V}_f$ as a separate pseudo-feature:
\begin{equation}
f^{(c)}(v) = \mathbb{I}[v = c],
\end{equation}%
where $\mathcal{V}_f$ is the set of all possible values of feature $f$. As a result, the values of a record $[v_1, \dots, v_F]$ consisting of both numerical and categorical values is transformed, after binning, into an expanded binary vector of pseudo-features:
\begin{equation}
\tilde{t} = [f_1^{(i_1)}(v_1), \dots, f_m^{(i_m)}(v_m), \dots, f_F^{(i_F)}(v_F)],
\end{equation}%
where each $f_m^{(i_m)}(v_m)$ indicates whether the value $v_m$ of feature $f_m$ activates the corresponding bin or category. This binarized representation enables the model to capture fine-grained, value-dependent interactions between features while maintaining sparsity.

Finally, we compute the mutual information between pseudo-features to quantify their statistical dependency. Given two pseudo-features $f_i^{(a)}$ and $f_j^{(b)}$, their mutual information is defined as:
\begin{equation}
\sum_{u \in \{0,1\}} \sum_{v \in \{0,1\}} 
P_{ij}(u, v) \log \frac{P_{ij}(u, v)}{P_i(u) P_j(v)},
\end{equation}%
where $P_{ij}(u, v)$ denotes the joint probability of activation values $u$ and $v$ for $f_i^{(a)}$ and $f_j^{(b)}$ respectively, while $P_i(u)$ and $P_j(v)$ denote the marginal probabilities. In practice, we estimate these probabilities empirically on the training split after pseudo-feature expansion. This makes the MI computation depend on binary pseudo-feature activations rather than raw numeric scales, while the capped bin count prevents excessively sparse or imbalanced pseudo-features from dominating the dependency graph. Appendix~\ref{sec:additional_abl} further shows that the downstream performance is stable across a broad range of MI thresholds, indicating that the guidance mechanism is not brittle to moderate estimation noise.

Based on the computed dependencies between pseudo-features, we can infer whether specific value ranges of the original features in dataset $\mathcal{T}$ exhibit statistical correlations. These correlations are then used to construct a dynamic feature dependency graph, which serves as a guiding structure to control the sampling process of $\theta$.

\subsubsection{Feature Selector.}
During sampling, we propose an explicit optimization strategy that filters out irrelevant feature-value pairs based on a static mutual information threshold. Specifically, for each target feature $f_{\text{target}}$, we define the set of relevant pseudo-features as:
\begin{multline}
\mathcal{R}(f_{\text{target}}) =
\{\, f_j^{(b)} \in \tilde{t}_{\text{prefix}} \;\mid\\
\operatorname{MI}\!\bigl(f_j^{(b)}, f_{\text{target}}\bigr) > \tau \,\}.
\end{multline}
where $\tilde{t}_{\text{prefix}}$ is the set of currently activated pseudo-features, and $\tau$ is a global threshold fixed before generation. By default we use a tuning-free setting and set $\tau$ to the median of MI values computed on the training set. This provides a robust scale for sparsification; the selected context remains instance- and step-dependent through $\mathcal{R}(f_{\text{target}})$.

Then, the generator $\theta$ conditions only on this relevant subset when generating the value for $f_{\text{target}}$:
\begin{equation}
P_\theta(v_{\text{target}} \mid \mathcal{R}(f_{\text{target}})).
\end{equation}%
This mechanism allows $\theta$ to dynamically adjust its attention over previously generated features for each generation step, thereby enabling fine-grained and value-sensitive dependency modeling.

\subsubsection{Logit Correction.}
While the selector module explicitly prunes irrelevant feature-value pairs, it also introduces a potential risk: the mutual information threshold $\tau$ is empirically set, and an overly strict threshold may suppress meaningful dependencies. As an alternative that avoids this issue, we propose an implicit logit correction mechanism.

Concretely, we compute the average mutual information between the current target feature $f_{\text{target}}$ and all previously generated pseudo-features in $\tilde{t}_{\text{prefix}}$:
\begin{equation}
\mu_{\text{sample}} = \frac{1}{|\tilde{t}_{\text{prefix}}|} \sum_{f_j^{(b)} \in \tilde{t}_{\text{prefix}}} 
\text{MI}(f_j^{(b)}, f_{\text{target}}).
\end{equation}%
We then compare this value with the dataset-wide average mutual information:
\begin{equation}
\Delta = \frac{\mu_{\text{sample}}}{\mu_{\text{train}}} - 1,
\end{equation}%
where $\mu_{\text{train}}$ denotes the expected mutual information between pseudo-features and $f_{\text{target}}$ across the training corpus.

We correct the final generation logit $z'$ for $f_{\text{target}}$:
\begin{equation}
z' = z \cdot (1 + \lambda \cdot \Delta).
\end{equation}%
Here $z$ is the unnormalised logit for the candidate value of $f_{\text{target}}$; all other logits are masked. The value $\lambda$ is a scaling hyperparameter. This correction sharpens the logit via softmax when the information provided by the prefix is highly relevant, i.e. $\Delta > 0$, and smooths it when the prefix contains little useful information, i.e. $\Delta < 0$~\citep{goodfellow2016deep}. Therefore, the model adaptively adjusts its generation confidence based on the information content of the already generated context.

\section{Experiment}

\begin{table*}[ht]
\centering
\footnotesize
\resizebox{0.95\textwidth}{!}{
\begin{tabular}{
ll
cccccccccccc
}

\toprule
&    & \multicolumn{2}{c}{Income} & \multicolumn{2}{c}{HELOC} & \multicolumn{2}{c}{Iris} & \multicolumn{2}{c}{Diabetes} & \multicolumn{2}{c}{MIC} & Housing\\

\cmidrule(lr){3-4} \cmidrule(lr){5-6} \cmidrule(lr){7-8} \cmidrule(lr){9-10} \cmidrule(lr){11-12} \cmidrule(lr){13-13}& & ACC $\uparrow$ & F1 $\uparrow$ & ACC $\uparrow$ & F1 $\uparrow$ & ACC $\uparrow$ & F1 $\uparrow$ & ACC $\uparrow$ & F1 $\uparrow$& ACC $\uparrow$ & F1 $\uparrow$ & \;MAPE $\downarrow$\\ 
\midrule

\multirow{2}{*}{Original} 
& DT & 83.54 & 0.73 & 67.90 & 0.68 & 100 & 1.00 & 82.73 & 0.33 & 96.76 & 0.96 & 0.27\\
& RF & 81.76 & 0.78 & 71.14 & 0.71 & 100 & 1.00 & 79.87 & 0.44 & 98.23 & 0.98 & 0.21\\
\midrule
\midrule

\multirow{2}{*}{TVAE} 
& DT & \underline{83.14} & \underline{0.73} & 64.70 & 0.63 & 55.17 & 0.45 & 81.93 & 0.37 & \textbf{96.76} & 0.95 & 0.37\\
& RF & 79.38 & 0.73 & 68.91 & 0.69 & 58.62 & 0.53 & 82.20 & \textbf{0.42} & \underline{96.76} & 0.96 & 0.30\\
\midrule

\multirow{2}{*}{CTGAN} 
& DT & 78.27 & 0.60 & 63.14 & 0.63 & 10.34 & 0.09 & \textbf{82.94} & 0.30 & 96.47 & 0.95 & 0.71\\
& RF & \underline{80.74} & 0.73 & 37.93 & 0.29 & 41.38 & \textbf{0.37} & 82.18 & 0.36 & 96.76 & 0.95 & 0.50\\
\midrule

\multirow{2}{*}{\textsc{TabSyn}} 
& DT & \textbf{83.32} & \textbf{0.75} & \underline{68.55} & 0.68 & 89.65 & 0.89 & 3.82 & 0.04 & \textbf{96.76} & 0.95 & \underline{0.32}\\
& RF & 79.61 & 0.59 & 70.48 & 0.70 & \textbf{100} & \textbf{1.00} & 8.14 & 0.12 & 96.76 & 0.95 & \textbf{0.23}\\
\midrule

\multirow{2}{*}{\textsc{GReaT}} 
& DT & 59.85 & 0.60 & 61.31 & 0.61 & 41.38 & 0.36 & \textbf{82.94} & 0.30 & 94.71 & 0.95 & 0.34 \\
& RF & 69.42 & 0.69 & 70.18 & 0.70 & 44.83 & 0.35 & 82.37 & 0.41 & 97.06 & \textbf{0.96} & 0.26\\
\midrule

\multirow{2}{*}{\textsc{GraDe}} 
& DT & 67.51 & 0.55 & 67.54 & 0.67 & \textbf{96.55} & \textbf{0.97} & \textbf{82.94} & \textbf{0.37} & 92.94 & 0.93 & \textbf{0.31}\\
& RF & 78.57 & 0.63 & \underline{70.73} & 0.71 & \textbf{100} & \textbf{1.00} & 81.86 & \textbf{0.42} & \underline{96.76} & 0.95 & \textbf{0.23}\\
\midrule
\multirow{2}{*}{\textsc{SPADA}} 
& DT & 77.65 & 0.50 & 61.62 & 0.61 & \textbf{96.55} & \textbf{0.97} & 81.17 & 0.32  & 96.17 & 0.95 & 0.40\\
& RF & \textbf{81.62} & 0.75 & 69.37 & 0.69 & \textbf{100} & \textbf{1.00} & 68.67 & 0.37  & \underline{97.06} & 0.96 & 0.25\\
\midrule
\midrule
\multirow{2}{*}{\textbf{Ours (w/FS)}} 
& DT & 80.66* & 0.68* & 68.45* & \underline{0.68*} & \textbf{96.55*} & \textbf{0.97*} & \textbf{82.94} & 0.30 & 96.47 & 0.95 & 0.34\\
& RF & 78.58* & \underline{0.75*} & \textbf{70.89} & \textbf{0.71} & \textbf{100*} & \textbf{1.00*} & 80.15 & 0.41 & 97.05 & 0.95 & 0.25\\
\midrule

\multirow{2}{*}{\textbf{Ours (w/LC)}} 
& DT & 82.58* & 0.72* & \textbf{69.52*} & \textbf{0.69*} & \textbf{96.55*} & \textbf{0.97*} & 75.58 & 0.34 & 96.47 & 0.95 & 0.64\\
& RF & 79.86* & \textbf{0.76*} & 70.58 & \underline{0.71} & \textbf{100*} & \textbf{1.00*} & 80.72 & 0.40 & \textbf{97.35} & \textbf{0.96} & 0.40\\

\bottomrule
\end{tabular}}
\caption{Performance of classifiers/regressors trained on synthetic data for downstream tasks. The best results are in \textbf{Bold}, and \underline{underline} indicates the second-best. ``Original'' denotes models trained and tested on the real dataset $\mathcal{T}$, while all other methods train on synthetic data and test on real data. ``ACC'' denotes accuracy and ``MAPE'' denotes Mean Absolute Percentage Error. Values marked with an asterisk (*) are statistically significantly different from \textsc{GReaT} (paired t-test, $p < 0.05$).}
\label{tab:downstream_utility}
\vspace{-10pt}
\end{table*}

\subsection{Datasets}

\textbf{Binary Classification.} The \emph{Adult Income} dataset \citep{adult_2} comprises 16 demographic and occupational variables and is used to predict whether an individual’s annual income exceeds a specified threshold. The \emph{Home Equity Line of Credit} (HELOC) dataset~\citep{heloc} includes 24 credit-related attributes extracted from credit reports, aiming to predict whether individuals will fully repay their HELOC balance within two years. To simulate a high-dimensional feature setting, we employ the \emph{Myocardial Infarction Complications} (MIC) dataset~\citep{heart}, which involves predicting myocardial infarction complications from 110 biological features measured on the first and third days of hospitalization.

\noindent\textbf{Multi-class Classification.} The \emph{Iris} dataset~\citep{iris_53} contains four numerical features measuring sepal and petal dimensions, aimed at classifying samples into iris species. The \emph{CDC Diabetes Health Indicators} dataset~\citep{burrows2017incidence} is a large-scale clinical study comprising over 250,000 records with 35 features, including demographics and laboratory results, designed to classify patients as healthy, Type 1 diabetes, or Type 2 diabetes.

\noindent\textbf{Regression.} The \emph{California Housing} dataset~\citep{house} comprises 10 variables describing housing and geographic attributes. The task is to predict the median value of owner-occupied homes.

\subsection{Evaluation Metrics}
\noindent\textbf{Downstream Utility.}  
To evaluate the downstream utility of the synthetic datasets, we generated synthetic data with the same sample size as the original datasets. We then trained decision tree (DT) and random forest (RF) models for both classification and regression tasks, following the intended task types associated with each dataset. These models represent a diverse range of learning paradigms to ensure comprehensive evaluation. For classification tasks, we reported \textit{accuracy} and \textit{F1 score}, while for regression tasks, we reported \textit{mean absolute percentage error (MAPE)}. The evaluation results are summarized in Table~\ref{tab:downstream_utility}.

\begin{figure*}[ht]
    \centering
    \includegraphics[width=\textwidth]{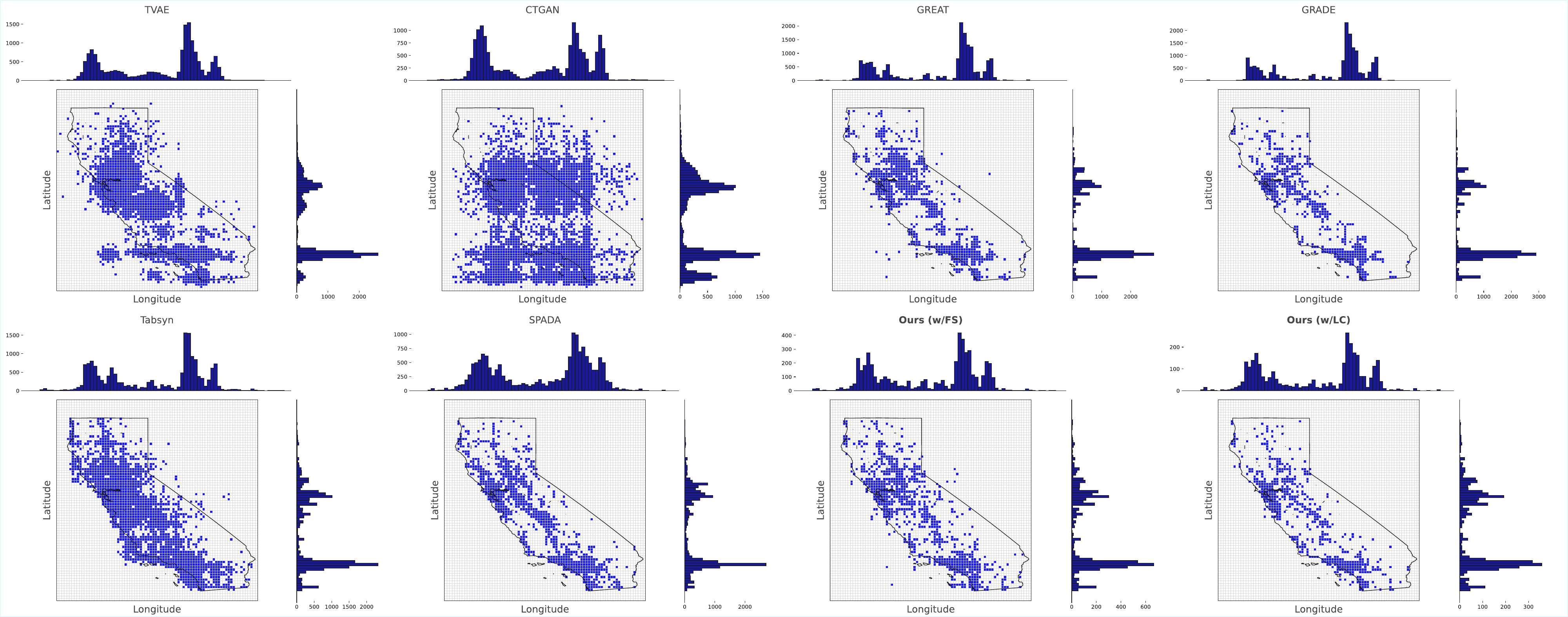}
    \caption{Comparison of the generated samples for the California Housing dataset, which includes characteristic information about various properties in California, USA. Joint histogram plots of the highly correlated variables Latitude and Longitude are shown. The black outline represents the true boundary of the state of California.}
    \label{fig:california_comparison}
    \vspace{-10pt}
\end{figure*}

\noindent\textbf{Data Fidelity.}  
Following~\citet{xu2025llmsbadsynthetictable}, we assess data fidelity with \textit{violation rates} under dataset-specific constraints. On California Housing, the violation rate is the probability that a generated data point lies outside the true geographical boundaries of California, providing a concrete measure of spatial constraint adherence. On Adult Income, we additionally evaluate a semantic consistency rule between \textit{education} and \textit{education-num}, where generated records that violate the canonical ordering of education levels and years of education are counted as invalid. This extends the evaluation beyond a single housing-boundary test. The Housing results are shown in Figure~\ref{fig:violation_rates}, and the multi-dataset constraint results are summarized in Appendix Table~\ref{tab:constraint_eval}. Additionally, we visualize the spatial distribution of the synthetic samples based on their geographic coordinates, as illustrated in Figure~\ref{fig:california_comparison}.

\noindent\textbf{Realism.}  
To measure the realism of the synthetic data, we trained a support vector machine (SVM) classifier~\citep{svm} using 5-fold cross-validation to distinguish between the original dataset and the synthetic dataset. The classification accuracy of this model serves as a proxy for realism: lower accuracy indicates that the synthetic data more closely resembles the real data and is therefore harder to distinguish. The corresponding results are reported in Appendix Table~\ref{tab:realism}.

\noindent\textbf{Privacy Protection.}  
In line with~\citep{zhang2023mixed}, we evaluated privacy protection by computing the \textit{Distance to Closest Record (DCR)} using the L1 norm~\citep{boyd2004convex}, measuring the proximity between synthetic samples and the nearest records in the original dataset. A \textit{higher DCR} implies reduced resemblance to any real individual and thus stronger privacy guarantees. Conversely, a \textit{lower DCR} reflects closer alignment with the real data distribution. The results are visualized in Appendix Figure \ref{fig:privacy_protection}.

\begin{figure}[ht]
    \centering
    \includegraphics[width=\linewidth]{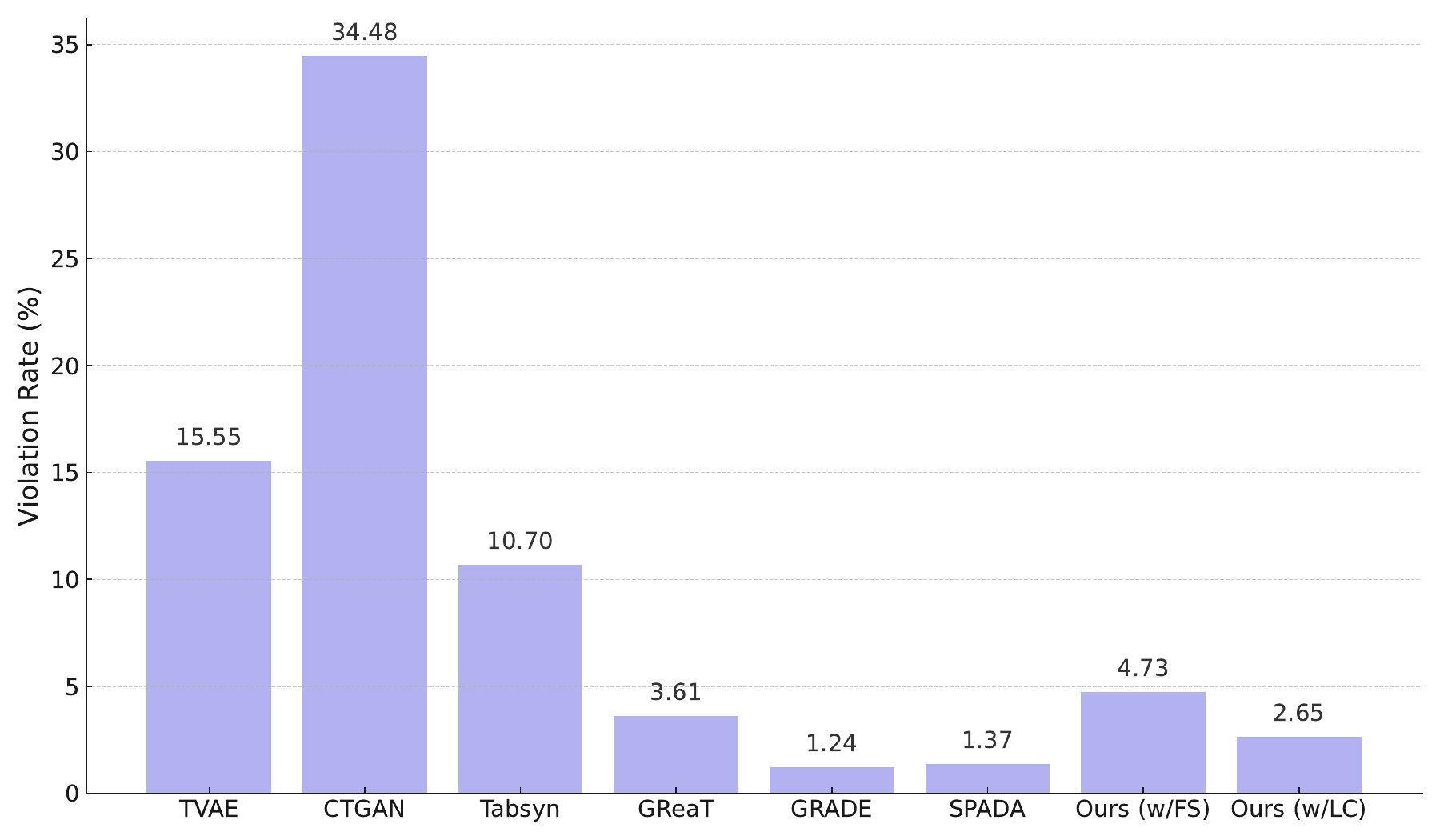}
    \caption{Violation rate, defined as the probability that a generated sample falls outside the true geographical boundaries of the state of California.}
    \label{fig:violation_rates}
    \vspace{-10pt}
\end{figure}

\subsection{Experimental Setup}
In all experiments, we use a batch size of 8, the AdamW optimizer~\citep{loshchilov2019decoupledweightdecayregularization}, and a learning rate of $1\mathrm{e}{-4}$. For sampling, we adopt nucleus sampling~\citep{Holtzman2020The} with $p=0.95$, a temperature of 1.0, and set the maximum generation length equal to the maximum sequence length observed in the training set. The mutual information threshold is empirically set to the median mutual information value computed from the training data.

\section{Result and Discussion}
\label{sec:result}
\noindent\textbf{(1) \underline{Downstream Utility}: Our method outperforms baselines, especially on small datasets.} 
In Table~\ref{tab:downstream_utility}, we show that \method consistently outperforms the baseline \textsc{GReaT} across almost all downstream tasks. The most significant improvement is seen on the \texttt{Adult} dataset, where our method achieves an increase of over 10 points in F1 score. Notably, \textsc{GReaT} performs poorly on smaller datasets such as \texttt{Iris}, where its accuracy reaches only 44.83\%, in stark contrast to our 96.55\%. We attribute this to the overfitting of \textsc{GReaT}, which only learns limited surface information of the training set and fails to capture meaningful patterns. In contrast, \method leverages mutual information to guide both the prefix construction and logits manipulation of the LLM, thereby shielding the model from being misled by superficial signals and resulting in improved robustness.

Compared to \textsc{TabSyn}, \method maintains more stable performance across datasets of varying sizes and dependency structures, demonstrating better generalization and robustness across diverse tabular domains. This stability is particularly evident in the consistent performance improvements across both classification and regression tasks.

\noindent\textbf{(2) \underline{Data Fidelity}: Both variants reduce violations, with stronger gains under different constraint types.}  
We observe that \method achieves a notable reduction in violation rates compared to all baseline methods on California Housing, as shown in Figure~\ref{fig:violation_rates}. Among our two variants, \textit{Logit Correction} achieves the largest improvement, reducing the violation rate by 1 point compared to \textsc{GReaT}, and achieving performance nearly on par with the GPT-4o-powered \textsc{SPADA}. Our \textit{Feature Selector}-based variant reduces the violation rate by approximately 6\% compared to \textsc{TabSyn}, demonstrating improved robustness and a superior ability to capture complex real-world data distributions.

As illustrated in the visualization in Figure~\ref{fig:california_comparison}, our method accurately captures the complex distribution of real-world data, with almost no synthetic samples falling outside the true boundaries of California. In contrast, \textsc{TVAE} and \textsc{CTGAN} struggle to learn such intricate spatial distribution patterns from limited training data and consequently fail to reproduce the correct spatial contours.

The additional semantic constraint evaluation on Adult Income further refines this picture. As shown in Appendix Table~\ref{tab:constraint_eval}, \textit{Feature Selector} reduces the education-consistency violation rate to 1.32\%, substantially below most baselines and much lower than \textit{Logit Correction}. This suggests that explicit context pruning is especially helpful when the target rule depends on a small set of semantically precise attributes, whereas implicit correction is more advantageous in the smoother spatial setting of Housing.

\noindent\textbf{Takeaways.} 
Compared to \textsc{GReaT} and \textsc{SPADA}, \method exhibits superior generalization on small-scale datasets such as \textit{Iris}, suggesting that mutual information-based guidance effectively mitigates overfitting to superficial token-level patterns—a limitation of autoregressive LLM generators. Dynamic adaptation of dependency structures allows \method to focus on relevant feature relationships rather than spurious correlations.

When comparing the two proposed guidance strategies, we observe distinct advantages for each approach. \textit{Logit Correction} achieves lower violation rates on Housing, indicating enhanced fidelity in spatially constrained domains where coherent relationships are crucial. \textit{Feature Selector} generally yields lower MAPE in regression tasks and more stable classification performance, likely due to its explicit filtering of spurious contextual signals. This explicit filtering is particularly beneficial in high-dimensional feature spaces where noise significantly impacts generation quality.

These results demonstrate that explicit and implicit guidance mechanisms provide complementary benefits depending on domain characteristics. \textit{Feature Selector} suits scenarios requiring aggressive noise filtering, while \textit{Logit Correction} better handles complex interdependencies. We note that on HELOC, \textit{Logit Correction} occasionally suppresses informative signals when contextual mutual information is underestimated, resulting in overly cautious generation. Addressing this through adaptive thresholding or hybrid strategies combining both approaches represents a promising direction for future work.

\section{Conclusion}
We introduce \method, a sparse and adaptive guidance framework for LLM-based tabular data generation that explicitly models the dynamic and sparse nature of feature dependencies. By discretizing features into pseudo-features and filtering context through mutual information, \method enables fine-grained and semantically accurate generation. Our method supports both explicit feature selection and implicit logit correction, offering flexible value-aware guidance during synthesis. 
Extensive experiments across six diverse datasets demonstrate that \method consistently improves generation quality, achieving up to +10.3\% F1 improvement over the LLM-based baseline \textsc{GReaT} on Adult and reducing policy violation rates by over 6\% on Housing. These results validate that value-sensitive dependency modeling leads to more realistic, controllable, and privacy-preserving synthetic tabular data.

\section*{Limitations}
While \method demonstrates significant improvements in tabular data generation, we acknowledge several avenues for future work:

\paragraph{(1) Modeling of Higher-Order Dependencies.}
Our guidance mechanism relies on pairwise mutual information to construct the dependency graph, which may not explicitly capture more complex, higher-order interactions where multiple features collectively influence a target. However, \method's autoregressive generation process partially mitigates this limitation. By conditioning each new value on the entire sequence of previously generated feature-value pairs, the underlying LLM can implicitly learn and leverage these multi-feature contexts during synthesis, moving beyond the purely pairwise signals used for guidance.

\paragraph{(2) Scalability of Preprocessing.}
For datasets with extremely high dimensionality, the one-time preprocessing step of computing the mutual information matrix could become computationally intensive. Nevertheless, this design choice was made deliberately to maximize efficiency during the critical generation phase. Since this computation is performed only once per dataset, the subsequent synthesis process is highly scalable. The sparse context provided by our \textit{Feature Selector} ensures that the inference cost remains low, avoiding the quadratic complexity of dense attention models at generation time.

\section*{Acknowledgments}
This work was partially supported by the Verband der Vereine Creditreform e.V.. 

\paragraph{Use of AI Assistants} The authors acknowledge the use of ChatGPT exclusively to refine the text in the final manuscript.


\bibliography{custom}

\appendix

\section{Appendix}
\label{sec:appendix}

\noindent The structure of Appendix is as follows:
\begin{itemize}
    \item Section \ref{sec:additional-the} provides theoretical foundations for \method, including information-theoretic justifications for the \textit{Feature Selector} and \textit{Logit Correction} mechanisms.
    \item Section \ref{sec:additional-opt} presents engineering optimizations that reduce computational overhead and improve generation reliability through supervised fine-tuning and rejection sampling strategies.
    \item Section \ref{sec:additional-exp} reports additional experimental results on distributional fidelity, realism and privacy preservation metrics, including density distribution analysis, discriminator accuracy and distance to closest record evaluation.
    \item Section \ref{sec:additional_abl} conducts an ablation study examining the impact of mutual information thresholds on downstream performance across different datasets.
    \item Section \ref{sec:additional_imp} details implementation specifics, including evaluation settings and comprehensive dataset statistics.
\end{itemize}

\section{Theoretical Foundations of \method}
\label{sec:additional-the}

In this section, we provide theoretical insights to motivate the design of our method \method, grounded in information theory and probabilistic modeling.

\subsection{Mutual Information as a Proxy for Dependency}
Let $X$ and $Y$ denote two (pseudo-)features from the transformed binary tabular space. The mutual information (MI) between $X$ and $Y$ is defined as:
\begin{equation}
    I(X; Y) = \sum_{x \in \mathcal{X}} \sum_{y \in \mathcal{Y}} P(x, y) \log \frac{P(x, y)}{P(x)P(y)}.
\end{equation}
This quantity measures the reduction in uncertainty of one variable given knowledge of the other. In our context, a high MI between a context feature and the target feature indicates a strong statistical dependency. Hence, selecting features with high $I(X; Y)$ for conditioning naturally enhances the relevance and coherence of the generated value.

\subsection{Feature Selector: An Information Bottleneck View}
The \textit{Feature Selector} aims to construct a reduced context $\mathcal{C} \subset \mathcal{F}_{\text{prefix}}$ such that:
\begin{equation}
    \mathcal{C} = \arg\max_{\mathcal{C} \subset \mathcal{F}_{\text{prefix}}} \sum_{f_i \in \mathcal{C}} MI(f_i; f_{\text{target}}), \quad \text{s.t. } |\mathcal{C}| \leq K,
\end{equation}
where $K$ is implicitly controlled by the mutual information threshold $\tau$. This formulation resembles the \textit{information bottleneck principle}, where one seeks to retain only the most informative subset of input variables for predicting the output while compressing irrelevant ones.

\subsection{Logit Correction: KL Divergence Justification}
Let $P^*(v \mid \mathcal{C})$ denote the optimal target value distribution given an ideal context $\mathcal{C}$, and let $P_\theta(v \mid \mathcal{C})$ be the model's predicted distribution. Minimizing the Kullback-Leibler divergence
\begin{equation}
    D_{\mathrm{KL}}(P^* \,||\, P_\theta) = \sum_{v} P^*(v \mid \mathcal{C}) \log \frac{P^*(v \mid \mathcal{C})}{P_\theta(v \mid \mathcal{C})}
\end{equation}
is equivalent to maximizing the log-likelihood under $P_\theta$. When $P_\theta$ underestimates the confidence due to weak context, we apply a logit rescaling based on mutual information strength:
\begin{equation}
    z'_v = z_v \cdot  (1 + \lambda \cdot \Delta), \quad \text{with } \Delta = \frac{\mu_{\text{sample}}}{\mu_{\text{train}}} - 1,
\end{equation}
where $z_v$ is the pre-softmax logit of value $v$. This can be interpreted as dynamically adjusting $P_\theta$ to better approximate $P^*$ when the context informativeness $\mu_{\text{sample}}$ deviates from expectation.

\subsection{Dynamic Dependencies and Conditional Relevance}
Finally, our binarization into value-aware pseudo-features enables approximation of conditional dependencies. For instance, mutual information between $f^{(a)}_i$ and $f^{(b)}_j$ (specific bins) reveals fine-grained value-conditioned patterns:
\begin{equation}
    I(f^{(a)}_i; f^{(b)}_j) \approx \mathbb{E}_{v_i, v_j \sim T} \left[\log \frac{P(v_i, v_j)}{P(v_i)P(v_j)}\right].
\end{equation}
This allows \method to dynamically adapt generation based on the evolving feature prefix, approximating conditional distributions without requiring explicit Bayesian graphs or rule sets.

It is worth noting that the dynamic nature of \method lies in its flexible selection of dependent features; the use of fixed hyperparameters does not affect the inherent dynamism of the method itself.

\textbf{Conclusion.} These theoretical foundations justify the design of both the Feature Selector and Logit Correction components of \method, offering principled mechanisms for sparse and adaptive control of the generation process.

\section{Additional Engineering Optimization}
\label{sec:additional-opt}

\begin{table*}[t]
\centering
\renewcommand{\arraystretch}{1.1}
\footnotesize
\resizebox{\textwidth}{!}{
\begin{tabular}{lcccccc|cc}
\toprule
Dataset & TVAE & CTGAN & \textsc{GReaT} & \textsc{TabSyn} & \textsc{GraDe} & \textsc{SPADA} & \textbf{Ours (w/FS)} & \textbf{Ours (w/LC)}  \\ \midrule
Income   & 80.23$\pm$0.01 & 74.08$\pm$0.02 & 99.95$\pm$0.00  & \textbf{54.75$\pm$0.02} & 99.98$\pm$0.00 & \underline{65.83$\pm$0.03}  & 73.80$\pm$0.01      & 88.28$\pm$0.01                         \\
HELOC    & 92.63$\pm$0.62 & 94.90$\pm$0.70 & 72.88$\pm$1.54 & \textbf{54.15$\pm$2.49} & 70.25$\pm$1.08 & 77.80$\pm$1.95 & \underline{60.50$\pm$2.49} & 95.55$\pm$1.27 \\
Iris     & 87.00$\pm$5.55 & 92.00$\pm$3.40 & 83.50$\pm$8.39 & \underline{53.00$\pm$4.04} & 57.00$\pm$9.28 & 67.50$\pm$5.81 & 54.50$\pm$8.89 & \textbf{51.50$\pm$3.45} \\
Diabetes & 84.58$\pm$1.31 & 75.52$\pm$1.38 & 66.83$\pm$1.58 & 99.81$\pm$0.08 & \textbf{64.82$\pm$0.63} & 99.70$\pm$0.26 & \underline{66.00$\pm$2.86} & 77.35$\pm$2.63 \\
Housing  & 74.30$\pm$0.33 & 89.48$\pm$0.92 & 66.22$\pm$2.08 & \textbf{50.28$\pm$0.95} & 66.51$\pm$1.20 & 69.33$\pm$2.34 & \underline{55.20$\pm$1.06} & 94.12$\pm$1.18 \\ \midrule
\rowcolor[rgb]{0.95,0.95,0.95}
Mean ($\downarrow$) & 83.35 & 85.20 & 77.88 & \textbf{53.55} & 71.31 & 76.83 & \underline{61.60} & 81.36 \\
\bottomrule
\end{tabular}}
\caption{Discriminator measure with a 5-fold cross-validation. Lower accuracy values indicate that the discriminator struggles to distinguish synthetic records from real data. \textbf{Bold} indicates the best performance, and \underline{underline} indicates the second-best.}
\label{tab:realism}
\end{table*}

\begin{figure*}[ht]
    \centering
    \includegraphics[width=\textwidth]{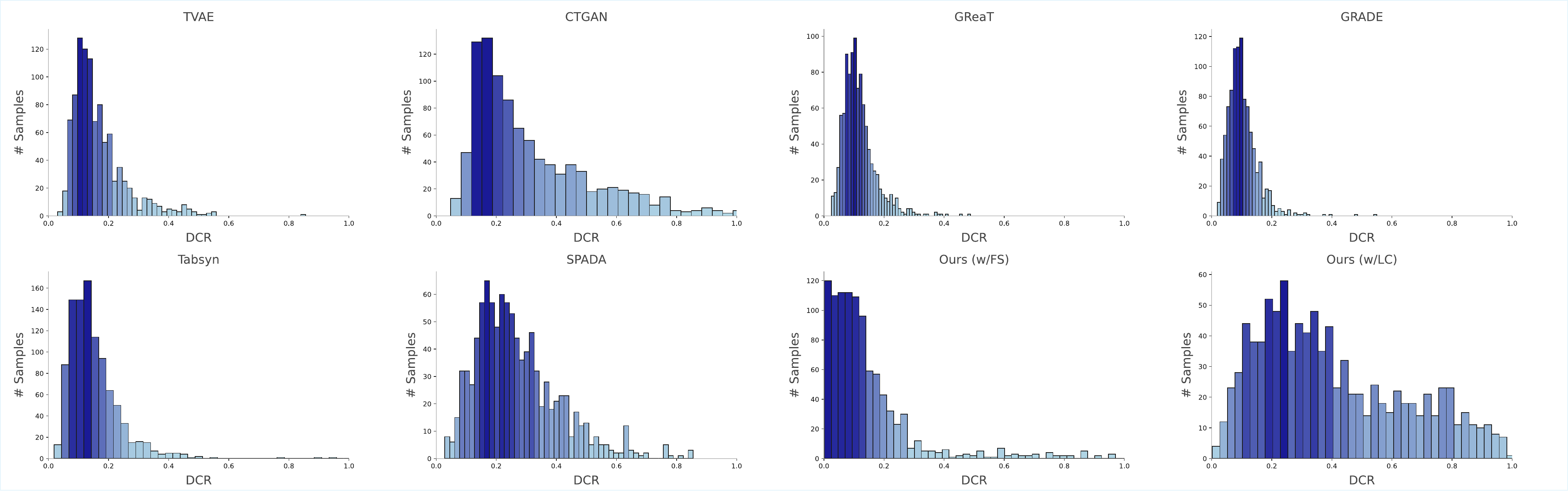}
    \caption{DCR for the California Housing dataset, evaluated with respect to the original training set. A lower DCR value demonstrates a high similarity between the synthetic data and the original data distribution, whereas a higher DCR value indicates enhanced privacy preservation in the synthetic data.}
    \label{fig:privacy_protection}
\end{figure*}

\subsection{Reduce Computational Overload}
In the original \textsc{GReaT} framework \citep{borisov2022language}, LLMs are trained to generate not only the feature values but also the template text (e.g., ``\texttt{feature is}''). We argue that this imposes unnecessary burden on the model and introduces redundant loss computations on tokens that are not informative for the target distribution.

To address this, we adopt a supervised fine-tuning (SFT) strategy that only computes loss over the tokens corresponding to feature values. During training, we mask out the template tokens and optimize the model solely on value tokens.

In inference, to remain consistent with the training objective, we select a random feature $f$ and prepend its corresponding template phrase, i.e., ``\texttt{feature is}'' as the prefix. The fine-tuned LLM is trained to fill in the missing value:
\begin{equation}
\hat{v}_f = \arg\max_{v_1, \dots, v_T} \; \prod_{t=1}^T P_\theta(v_t \mid v_{<t}, \text{``}f \text{ is''}).
\end{equation}%
This design reduces the effective token-level computation by approximately $\sim 75\%$, as the model bypasses learning fixed template components. As a result, \method achieves significantly faster training and inference compared to \textsc{GReaT}, while maintaining comparable or better generation quality (see Section~\ref{sec:result} for empirical results).

\subsection{Rejection Sampling}
Due to the inherent degeneration in text generation, LLM-based tabular data generators inevitably produce illegal values occasionally. For instance, a model may generate non-numeric strings for numeric variables, or generate out-of-domain tokens for categorical variables, i.e., tokens that do not appear in the training data for a given feature.

The \textsc{GReaT} framework applies heuristic rules to remove any generated feature values that contain invalid tokens. However, this approach not only leads to a loss of information from the original sample, but also introduces null-value noise into the dataset, which may negatively affect downstream learning tasks.

As a more efficient alternative, we adopt a rejection sampling strategy by explicitly constraining the output token space during decoding. Specifically, we restrict the model’s vocabulary such that only tokens within the predefined valid range for each feature type can be sampled. 

\section{Additional Experimental Results}
\label{sec:additional-exp}
\noindent\textbf{Realism.}
To further demonstrate the effectiveness of \method, in Table~\ref{tab:realism}, we report the discriminator measure with a 5-fold cross-validation. We also visualize the Distance to Closest Record (DCR) distributions for the California Housing dataset with respect to the original training set in Figure~\ref{fig:privacy_protection}.

\paragraph{Constraint Fidelity Beyond Housing.}
Table~\ref{tab:constraint_eval} extends the violation-rate evaluation to two different kinds of constraints: a spatial boundary rule on California Housing and a semantic consistency rule on Adult Income. The results show that \textit{Feature Selector} transfers particularly well to the semantic rule, whereas \textit{Logit Correction} remains strongest on the spatial constraint, reinforcing that the two guidance strategies are complementary.

\begin{table*}[ht]
\centering
\footnotesize
\resizebox{0.95\textwidth}{!}{
\begin{tabular}{lccccccc}
\toprule
Dataset & TVAE & CTGAN & \textsc{GReaT} & \textsc{TabSyn} & \textsc{SPADA} & \textbf{Ours (w/FS)} & \textbf{Ours (w/LC)} \\ \midrule
Income  & 4.21$\pm$0.64 & 34.41$\pm$1.18 & \textbf{0.00$\pm$0.00} & 2.32$\pm$0.49 & 3.59$\pm$0.98 & \underline{1.32$\pm$1.16} & 16.47$\pm$4.51 \\
Housing & 15.55$\pm$0.55 & 34.48$\pm$0.72 & 3.61$\pm$0.28 & 10.70$\pm$0.72 & 5.26$\pm$0.62 & 4.73$\pm$0.59 & \textbf{2.65$\pm$0.59} \\ \midrule
Mean ($\downarrow$) & 9.88 & 34.40 & \textbf{1.81} & 6.51 & 4.41 & \underline{3.03} & 9.56 \\
\bottomrule
\end{tabular}}
\caption{Violation rates under two dataset-specific constraints. For Adult Income, we measure the consistency between \textit{education} and \textit{education-num}; for California Housing, we measure whether generated coordinates fall outside the California boundary. Lower is better.}
\label{tab:constraint_eval}
\end{table*}

\begin{figure*}[t]
    \centering
    \begin{subfigure}[b]{0.48\linewidth}
        \centering
        \includegraphics[width=\linewidth]{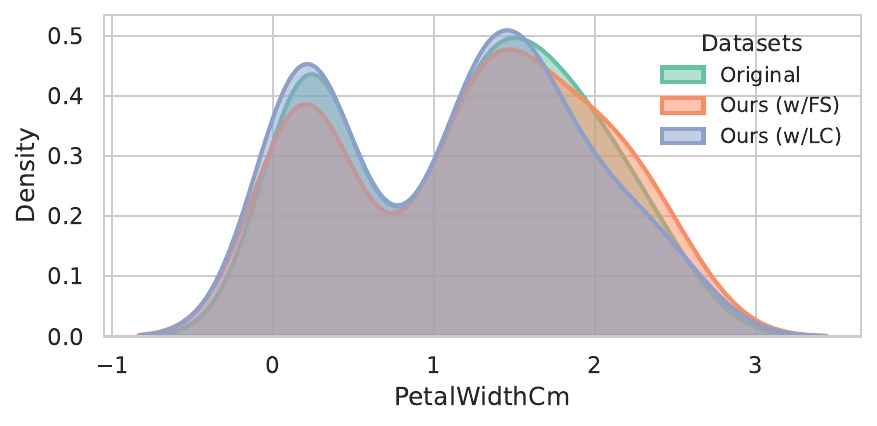}
    \end{subfigure}
    \begin{subfigure}[b]{0.48\linewidth}
        \centering
        \includegraphics[width=\linewidth]{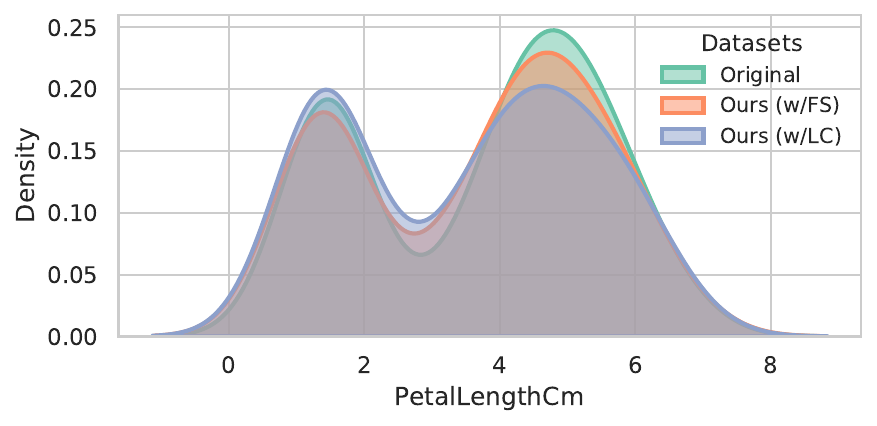}
    \end{subfigure}
    
        \begin{subfigure}[b]{0.48\linewidth}
        \centering
        \includegraphics[width=\linewidth]{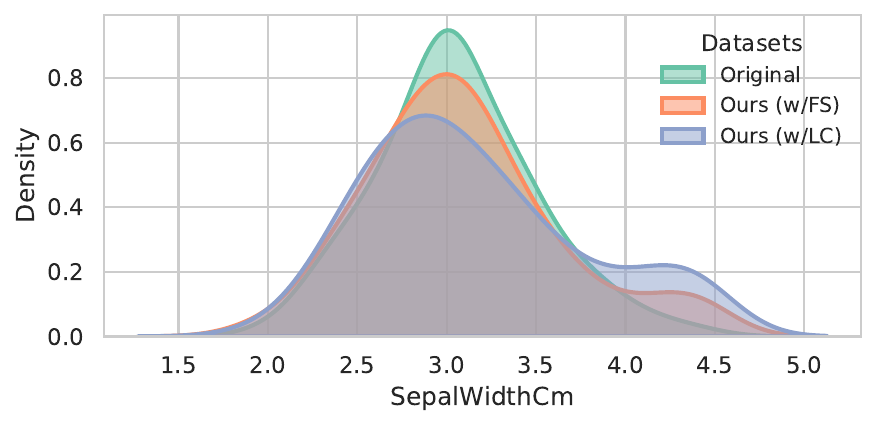}
    \end{subfigure}
    \begin{subfigure}[b]{0.48\linewidth}
        \centering
        \includegraphics[width=\linewidth]{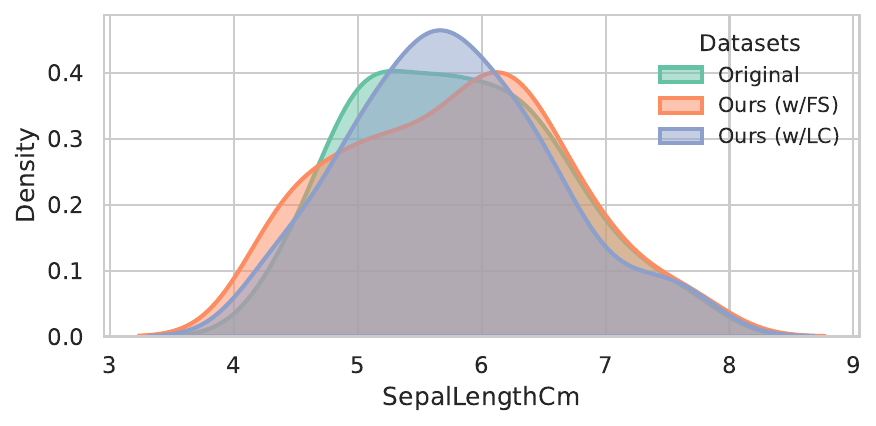}
    \end{subfigure}
    \caption{Visualization of the density distributions of Sepal and Petal lengths and widths on the Iris dataset, comparing the original and synthetic data.}
    \label{fig:iris}
\end{figure*}

\paragraph{Distributional Fidelity.} 
To further validate the distributional quality of our synthetic data, we examine the density distributions of all four numerical features in the Iris dataset. Figure~\ref{fig:iris} compares the original data with outputs from both \method variants. The results demonstrate that \textit{Feature Selector} and \textit{Logit Correction} effectively capture the underlying patterns, with synthetic distributions closely matching the characteristic shapes and ranges of sepal and petal measurements. This visual evidence complements our quantitative metrics and confirms that \method maintains high fidelity across different feature types and scales.

\begin{table*}[ht]
\centering
\footnotesize
\begin{tabular}{ll|c|cc}
\toprule
Dataset                  &               & \textsc{GReaT} & \textbf{Ours (w/FS)}    & \textbf{Ours (w/LC)}     \\ \midrule
\multirow{2}{*}{Income}  & Training & 6 h 10 min             & 15 min             & 15 min          \\
                         & Sampling & 9 sec                     & 0.2 sec           & 0.2 sec \\ \midrule
\multirow{2}{*}{HELOC}   & Training & 1 h 47 min            & 1 h 12 min            & 1 h 12 min         \\
                         & Sampling & 45 sec                    & 0.5 sec           & 0.5 sec  \\ \midrule
\multirow{2}{*}{Iris}    & Training & 17 sec                    & 20 sec              & 20 sec   \\
                         & Sampling & 4 sec                     & 0.07 sec &  0.07 sec\\ \midrule
\multirow{2}{*}{Housing} & Training & 1 h 18 min             & 52 min              & 52 min     \\
                         & Sampling & 8 sec                     & 0.4 sec           & 0.4 sec \\ \bottomrule
\end{tabular}
\caption{Average end-to-end training time, including one-time preprocessing for pseudo-feature construction and MI estimation, and sampling time per instance.}
\label{tab:time_cost}
\end{table*}

\paragraph{Time Cost.}
Table~\ref{tab:time_cost} summarizes the training durations and average per-sample sampling times for both the baseline methods and our proposed models.

\begin{figure*}[t]
    \centering
    \begin{subfigure}[b]{0.48\linewidth}
        \centering
        \includegraphics[width=\linewidth]{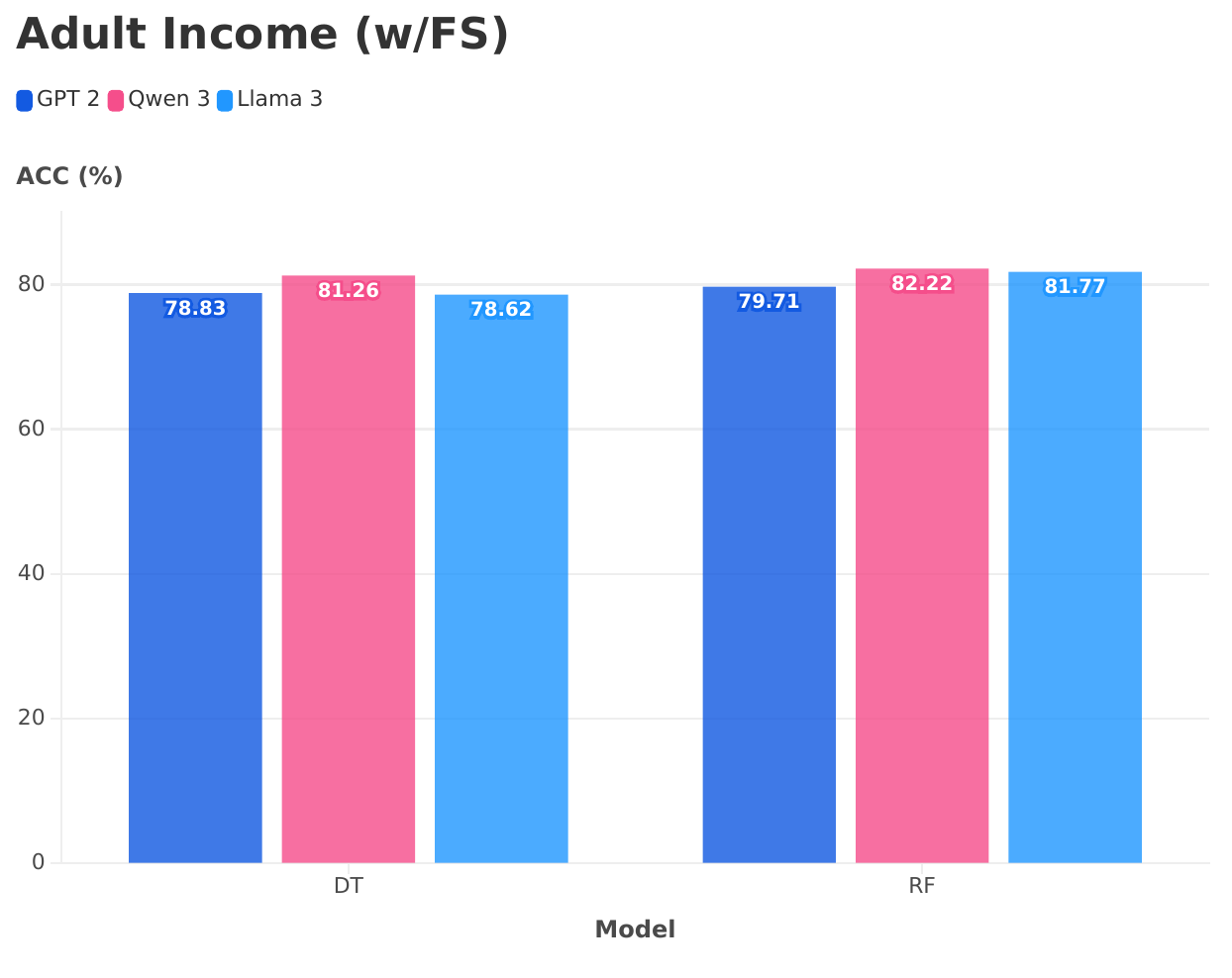}
    \end{subfigure}
    \begin{subfigure}[b]{0.48\linewidth}
        \centering
        \includegraphics[width=\linewidth]{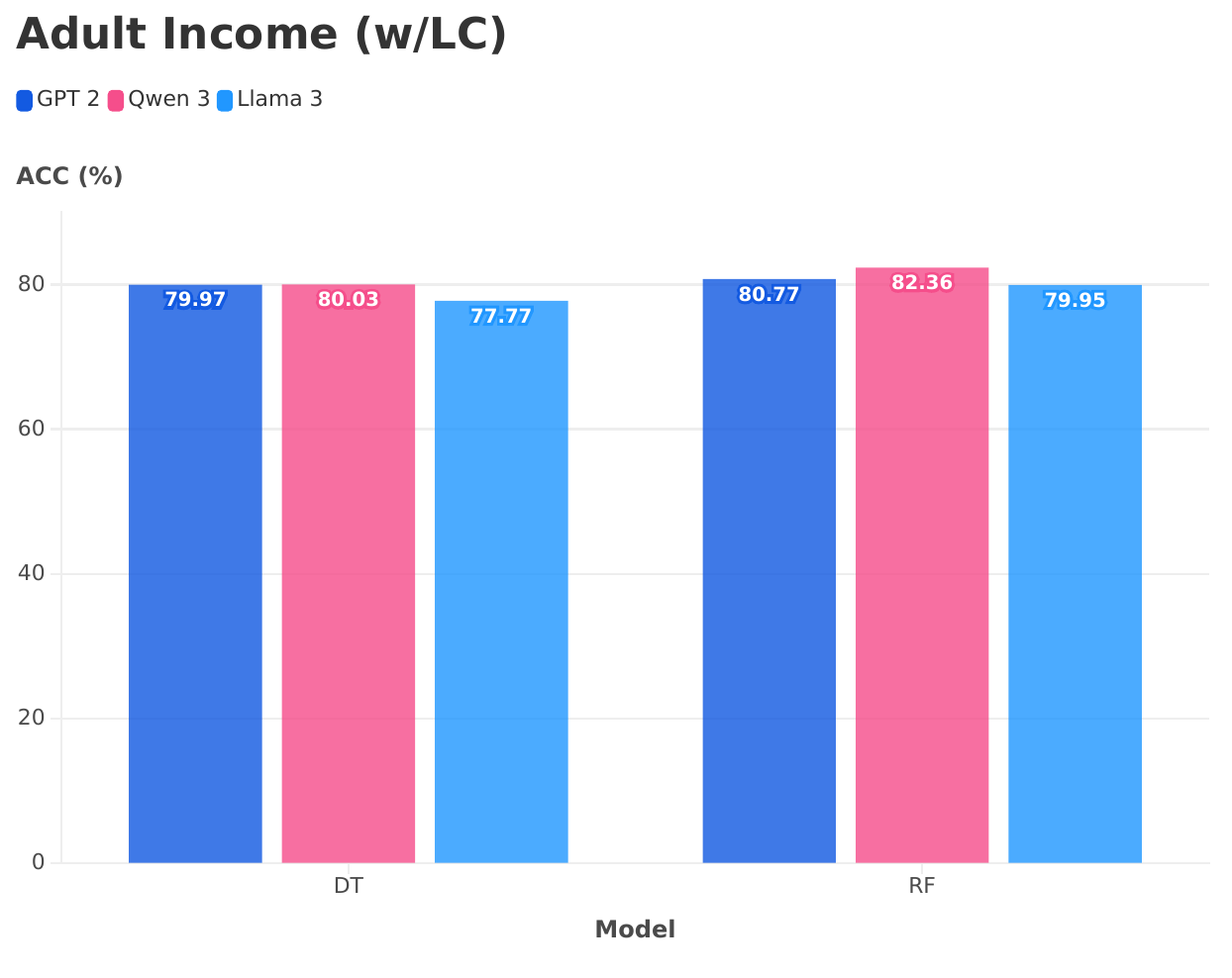}
    \end{subfigure}
    
        \begin{subfigure}[b]{0.48\linewidth}
        \centering
        \includegraphics[width=\linewidth]{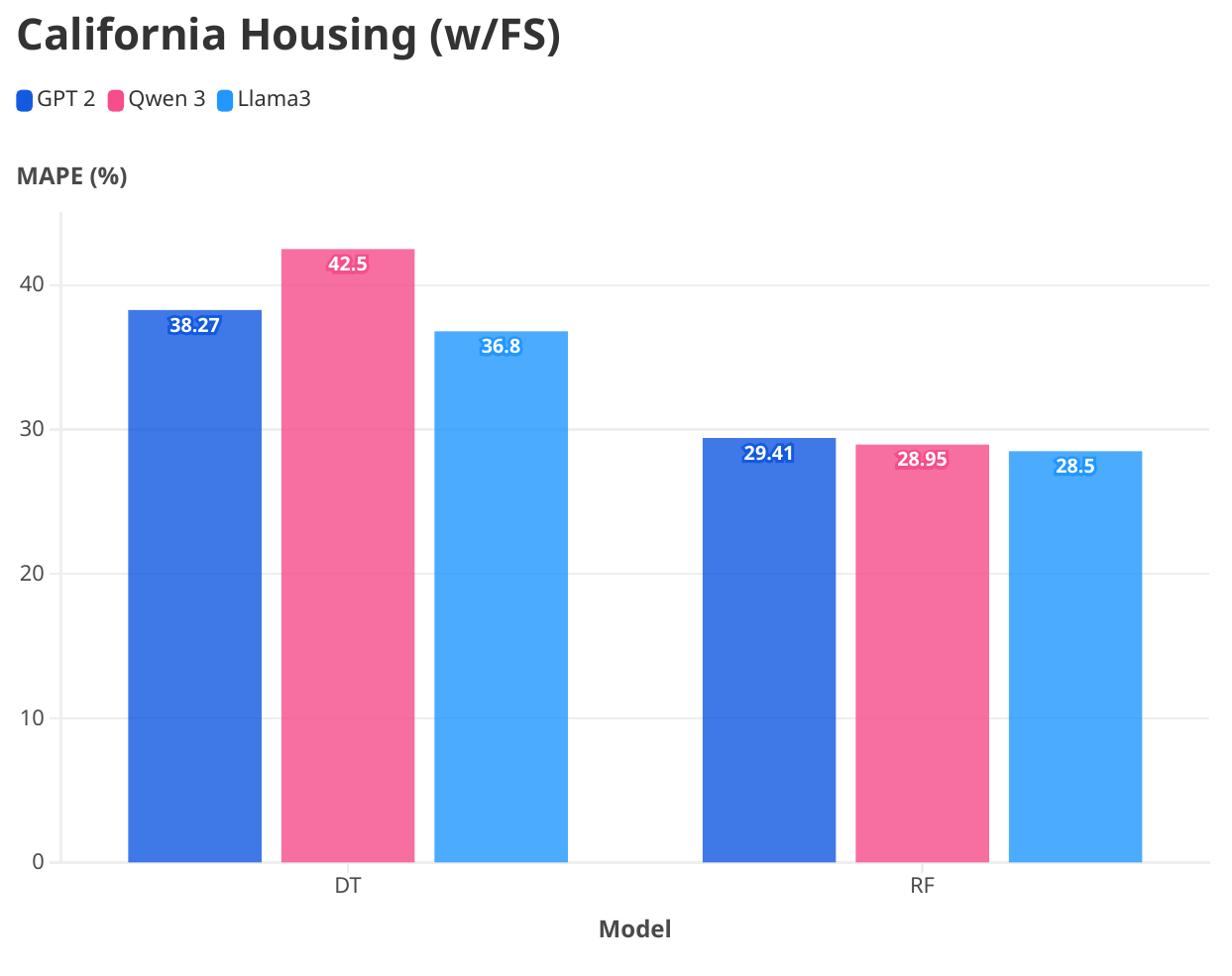}
    \end{subfigure}
    \begin{subfigure}[b]{0.48\linewidth}
        \centering
        \includegraphics[width=\linewidth]{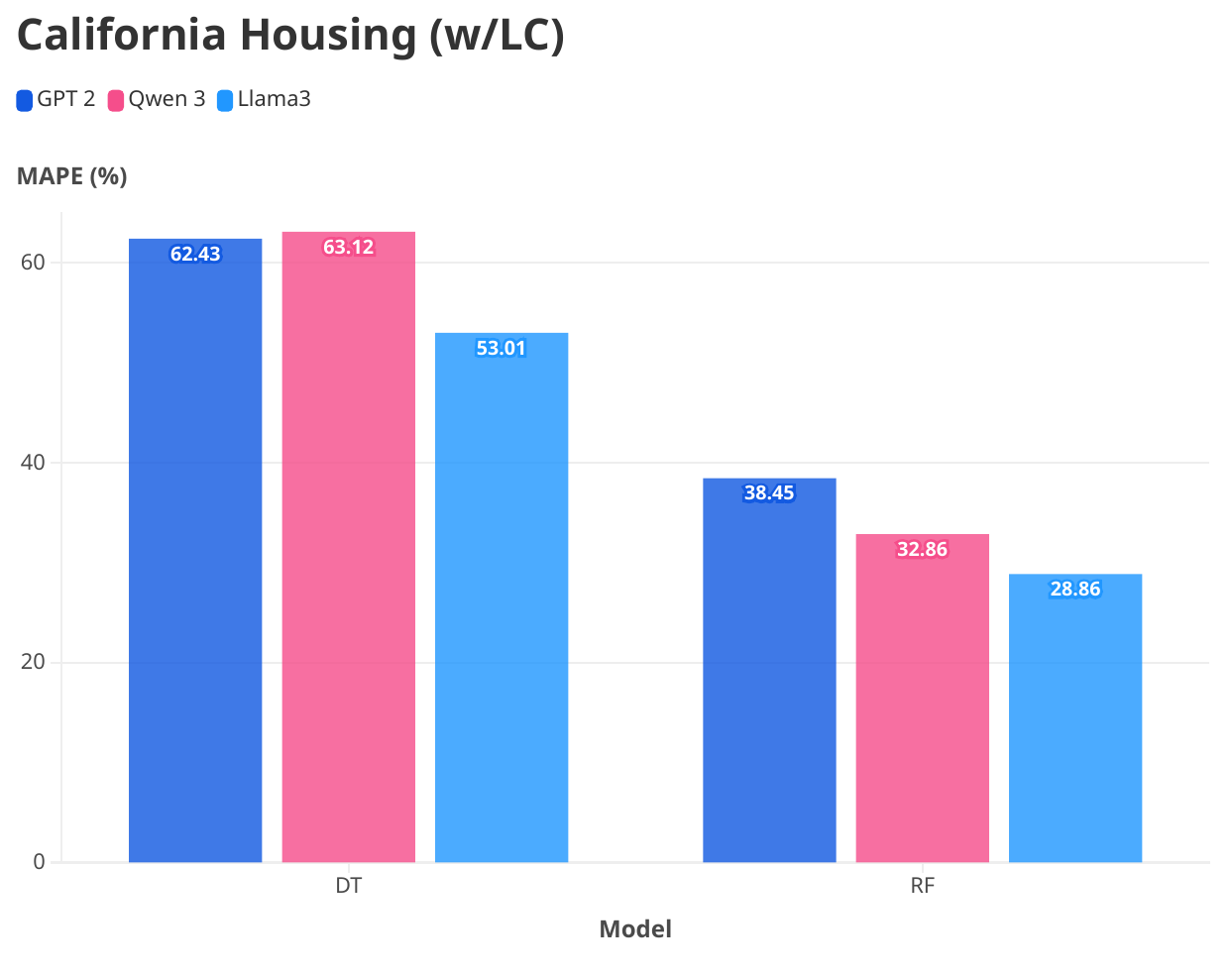}
    \end{subfigure}

    \caption{The performance of \method with different LLMs on classification and regression tasks.}
    \label{fig:diff_LMs}
\end{figure*}

\paragraph{Performance Across Different LLMs.} 
To demonstrate the robustness of \method across different model architectures, we evaluate both variants using GPT-2, Qwen-3, and Llama-3 as base models. Figure~\ref{fig:diff_LMs} presents results on Adult Income and California Housing datasets. Across all tested architectures, \method maintains consistent performance patterns, with Llama-3 generally achieving the best results. 

Notably, the relative advantages of \textit{Feature Selector} and \textit{Logit Correction} remain stable across different models, indicating that our method's effectiveness is architecture-agnostic rather than dependent on specific LLM characteristics.

\section{Ablation Study: Impact of MI Thresholds}
\label{sec:additional_abl}
\begin{figure*}[t]
    \centering
    \begin{subfigure}[b]{0.49\linewidth}
        \centering
        \includegraphics[width=\linewidth]{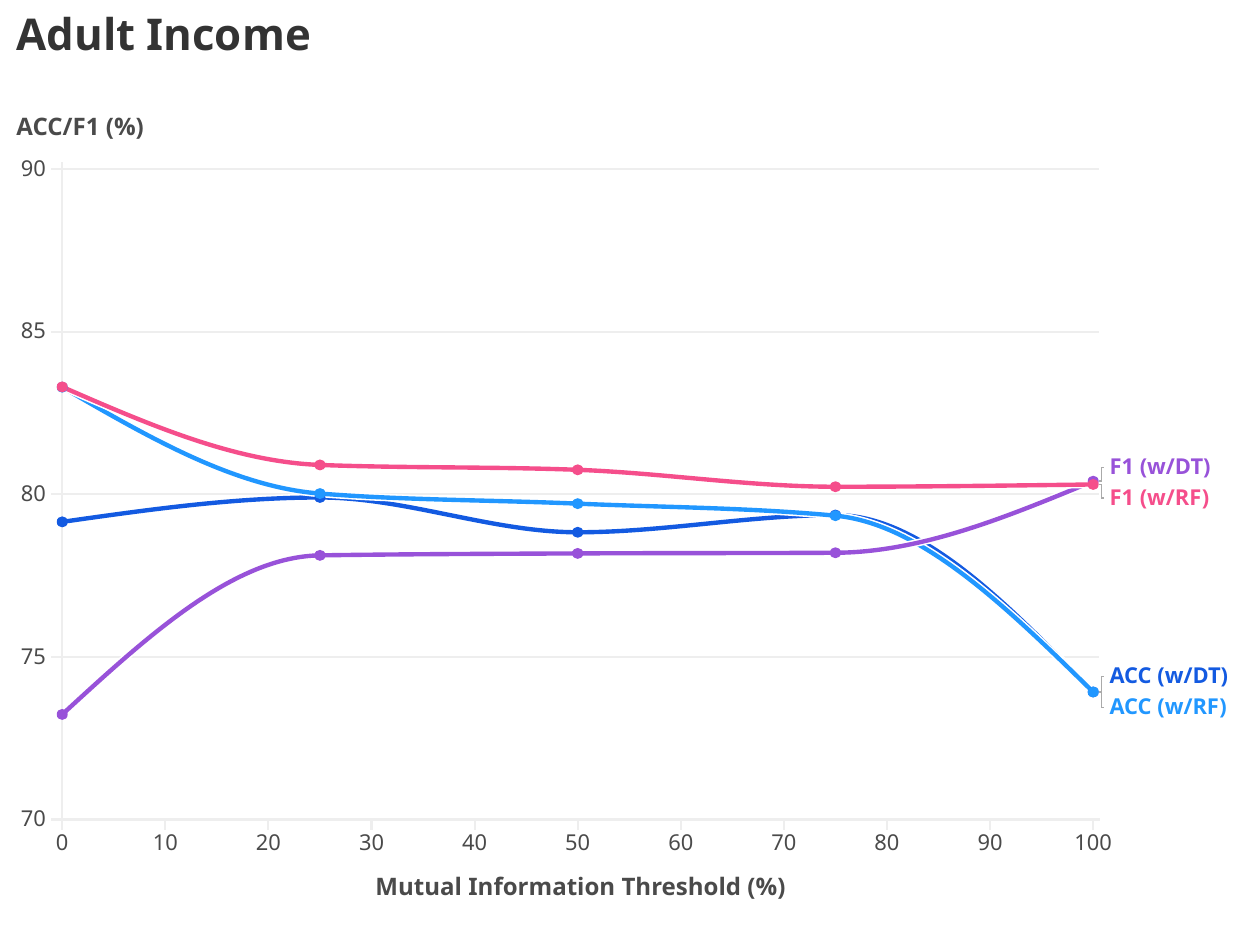}
    \end{subfigure}
    \begin{subfigure}[b]{0.49\linewidth}
        \centering
        \includegraphics[width=\linewidth]{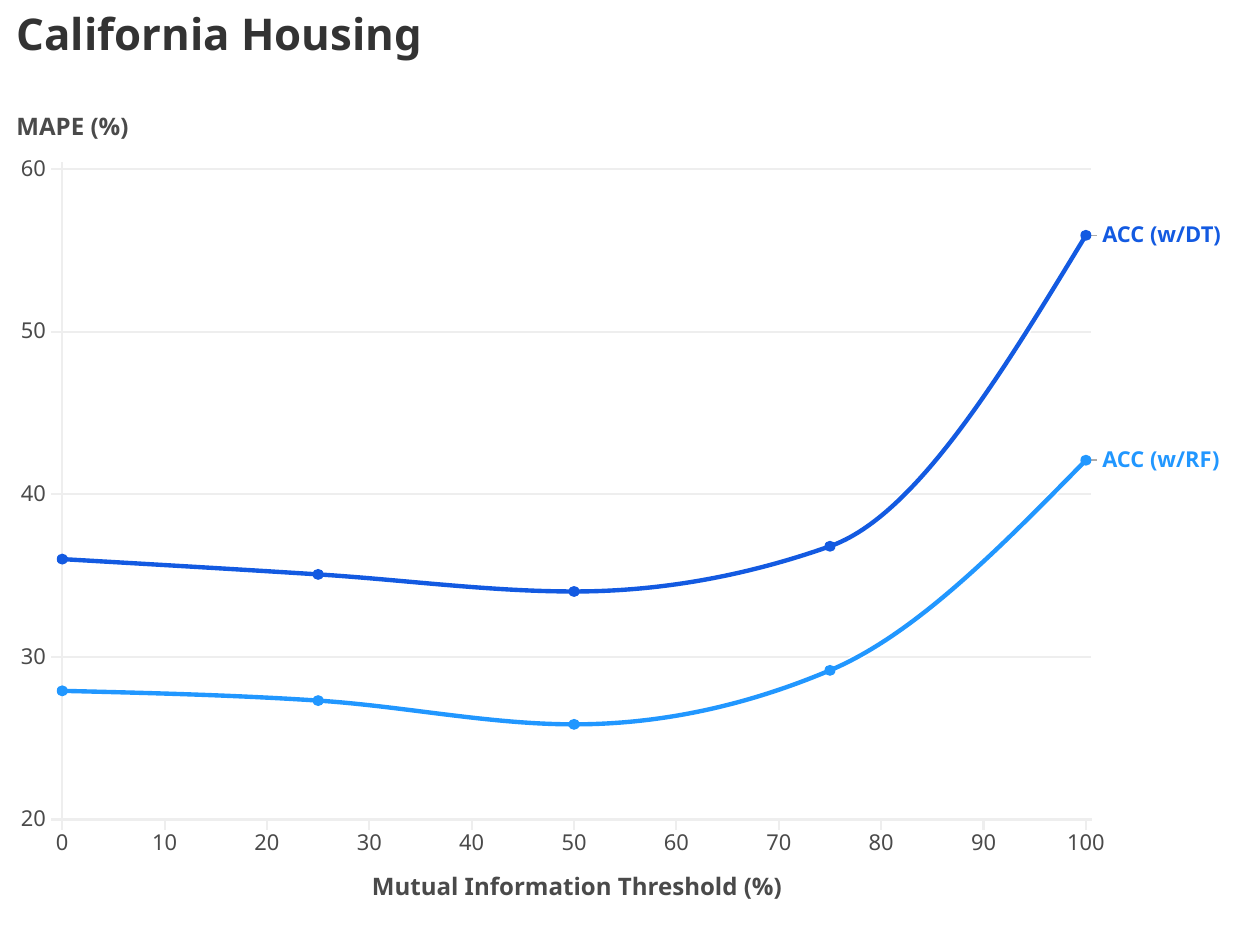}
    \end{subfigure}
    \caption{Impact of MI thresholds on downstream performance. The MI threshold refers to the proportion of training samples with the lowest MI; for instance, a 25\% threshold considers the bottom 25\% of samples sorted in ascending order by mutual information.}
    \label{fig:MI_threshold}
\end{figure*}

We examine how the MI threshold $\tau$ affects downstream performance by testing different threshold values on Adult Income and California Housing datasets. Figure \ref{fig:MI_threshold} shows distinct patterns across the two datasets. 

On Adult Income, performance remains relatively stable across most threshold ranges, with accuracy and F1 scores varying within 5 points. This stability suggests the dataset contains many genuinely irrelevant feature dependencies that can be safely filtered. California Housing shows different behavior. Performance stays stable until around 60-70\% threshold, then degrades sharply. MAPE jumps from $\sim$ 25\% to over 50\%, indicating that aggressive pruning removes important dependencies in this spatially-constrained domain.

These results reflect the datasets' inherent characteristics. Adult Income contains demographic features with clear independence, making sparse modeling effective. Housing data involves interconnected geographic and economic variables that require more careful dependency preservation. We set $\tau$ to the median MI value from training data (typically around 50\%) as a reasonable balance between sparsity and information retention.

\section{Implementation Details}
\label{sec:additional_imp}

\subsection{Evaluation Settings}
For the machine learning efficiency and discriminator experiments, we additionally use decision tree (DT), random forest (RF), linear/logistic regression (LR) and support vector machine (SVM) models from the Scikit-Learn package \citep{sklearn_api}.

\subsection{Datasets}
For all datasets, we use an 80\%/20\% train-test split for model training and evaluation. Table \ref{tab:datasets} provides comprehensive statistics for these datasets. 

\begin{table*}[ht] 
    \centering
    \footnotesize
    \begin{tabular}{lccccccc}
            \toprule
            Dataset  & Domain & \# Samples & \# Features & Task   & \# Classes\\
            \midrule
            Income \citep{adult_2}  & Social     & 48,842   & 15     & Classification    & 2  \\
            HELOC \citep{heloc}  & Finance     & 10,459     & 24             & Classification & 2\\
            Iris \citep{iris_53}   & Biology      & 150   & 5         & Classification & 3 \\
            Diabetes \citep{burrows2017incidence}  & Healthcare & 253,680   & 20        & Classification     & 3  \\
            MIC \citep{heart}  & Biology & 1,360   & 111        & Classification     & 2 \\
            Housing \citep{house} & Real Estate     & 20,640      & 10           & Regression & - \\
        \bottomrule
        \end{tabular}
  \caption{The statistics of the datasets employed in our experiments. \# Samples, \# Features and \# Classes denote the numbers of samples, features and classes in tabular datasets, respectively.} 
  \label{tab:datasets}
\end{table*}

\subsection{Hyperparameter}
To ensure the reproducibility of our reported experimental results, we list the hyperparameters used for each dataset in Tab.~\ref{tab:paras}. Furthermore, we employed nucleus sampling with $p = 0.7$ and temperature $= 1$ during decoding.

\begin{table*}[ht]
\centering
\footnotesize
\begin{tabular}{lcccccc}
\toprule
\multicolumn{1}{c}{} & Income & HELOC  & Iris   & Diabetes & MIC    & Housing \\ \midrule
Bins                 & 5      & 10     & 5      & 20       & 20     & 10      \\
MI threshold         & 0.0004 & 0.0142 & 0.1190 & 0.0063   & 0.0004 & 0.004   \\
$\lambda$            & 1.0    & 1.0    & 1.2    & 0.8      & 0.8    & 1.0     \\ 
\bottomrule
\end{tabular}
\caption{The hyperparameters used in our experiments. We performed a grid search over $\lambda$ in the range of 0.8 to 1.2 with a step size of 0.2, and over the number of bins in the range of 5 to 20 with a step size of 5.
}
  \label{tab:paras}
\end{table*}

\end{document}